\documentclass{article}

\usepackage[preprint]{nips_2018}

\usepackage[utf8]{inputenc} 
\usepackage[T1]{fontenc}    
\usepackage{url}            
\usepackage{booktabs}       
\usepackage{amsfonts}       
\usepackage{nicefrac}       
\usepackage{microtype}      
\usepackage{graphicx}

\usepackage{listings}
\usepackage{color}

\definecolor{gray}{rgb}{0.5,0.5,0.5}
\definecolor{myyellow}{RGB}{221, 187, 17}
\definecolor{myblue}{RGB}{52, 167, 255}
\definecolor{orange}{RGB}{255, 165, 0}

\usepackage{enumitem}\setlist[description]{font=\textendash\enskip\scshape\bfseries}
\usepackage{blindtext}
\usepackage{floatrow}
\newfloatcommand{capbtabbox}{table}[][\FBwidth]

\begin{document}
\title{Reimplementation and Reinterpretation of the Copycat Project}

\author{
Hongyi Huang \thanks{Currently in Viewpoint School; 23620 Mulholland Hwy,
Calabasas, CA 91302. Alternative email address: f.huang19@viewpoint.org.}\\
Northeastern University \\
440 Huntington Ave \\
Boston, MA 02115 \\
\texttt{h.hongyi@northeastern.edu}
}



\maketitle

\lstset{
  aboveskip=1.3 \medskipamount,
  belowskip=1 \medskipamount
}

\begin{abstract}
We present the reinterpreted and reimplemented \textit{Copycat} project, an architecture solving letter analogy domain problems. To support a flexible implementation change and rigor testing process, we propose a implementation method in DrRacket by using functional abstraction, naming system, initialization, and structural reference. Finally, benefits and limitations are analyzed for cognitive architectures along the lines of \textit{Copycat}.
\end{abstract}

\section{Introduction}
Copycat intends to model and prove that analogies are the source of human behavioral fluidity and creativity~\cite{Mitchell1993Analogy-makingPerception}. These letter analogy problems are given for the program to solve. For example, the problem abc : abd :: ijk : ? could have answers ijl, ijd, or even abd --- L being the successor of K, D being the literal letter substitution, and \textit{abd} being the complete direct substitution. However, humans give the answer \textit{ijl} more often than other answers statistically. Other answers are the result of different mental pressures producing different plausible answers. With this in mind, the model should produce answers in similar frequency. 
Formally, the letter analogy problem is denoted as initial : modified :: target : answer. The interpretation for Copycat is written in 1980s, before the bloom of reinforcement learning and machine learning. On one hand, these two field could potentially formalize the analogy problem. On the other hand, traditional reinforcement learning models has yet to address the viability to act and learn broadly and fluidly. This paper tries to connect each method by analyzing their respective benefits and limitations. 
\begin{figure}[h]
\centering
    \includegraphics[height=1.7in, width=2.8in]{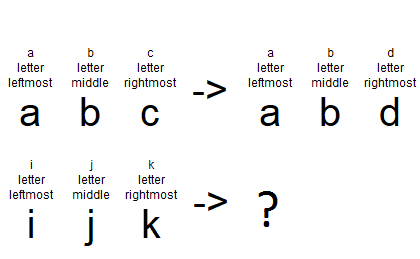}
    \caption{The initial configuration of Copycat's \textit{Workspace} with descriptions attached.}
\label{workspaceInit}
\end{figure}

\section{Background}
\subsection{Research Approach}
Cognitive architectures are aspects of cognition that remain constant in an agent~\cite{Newell1990UnifiedCognition}. Copycat is an example of so that consists of three main constant components to solve letter-domain analogy problems --- \textit{Workspace}, \textit{Coderack}, and \textit{Slipnet}~\cite{Mitchell1993Analogy-makingPerception,Hofstadter1995TheAnalogy-making}. The architecture is a step towards Newell and Minsky's urge to both bridge the symbolism-connectionism gap and to address all different layers of abstraction in human actions~\cite{Minsky1986TheMind, Newell1990UnifiedCognition}. Research in cognitive architectures involves dealing with unknown components of cognition, thus traditional Popperian data-hypothesis falsification process does not apply~\cite{Newell1990UnifiedCognition, Cooper2007TheAnalysis}. In the experimental process of these types of studies, some long-held assumptions are assumed to be true in order to test another central hypothesis. Copycat's hypothesis is that analogies are the source of human behavioral fluidity and creativity. To evaluate progress scientifically, Lakatos proposed that these research programs are composed of assumptions in a \textit{core} and hypothesis in a \textit{protective belt}, in which the assumptions within the belt is adjusted when empirical evidence refutes the hypothesis~\cite{Lakatos1976FalsificationProgrammes}. This standard has been commonly used to assess how much progress has been made for research in other modern cognitive architectures~\cite{Schultheis2009ComputationACT-R}. To make scientific progress, later research projects must change and manipulate parts of the earlier architectures, such as the Marshall's \textit{Metacat} that builds upon Copycat's architecture~\cite{Marshall1999Metacat:Perception}. If such architectures are easy to implement, test and understand, this would not be much of a problem at practice. However, these architectures are often complex and include handcrafted knowledge to hold the \textit{core} assumption of cognition constant. Later researchers could spend a significant amount of time to investigate software not maintained or documented properly. In this case, Marshall had to rewrite the code in Chez Scheme for the Metacat project instead of reusing Common-Lisp code from a few years ago. Maintaining a consistent and efficient practice would contribute to reusable code.

\begin{figure}[h]
\begin{floatrow}
\ffigbox{%
    \includegraphics[height=1.2in, width=1.3in]{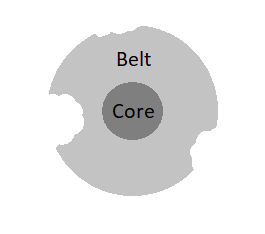}
}{%
  \caption{Lakatosian's core assumptions and falsifiable protective belts when facing inconsistency.}
}
\ffigbox{%
    \includegraphics[height=1.2in, width=2.5in]{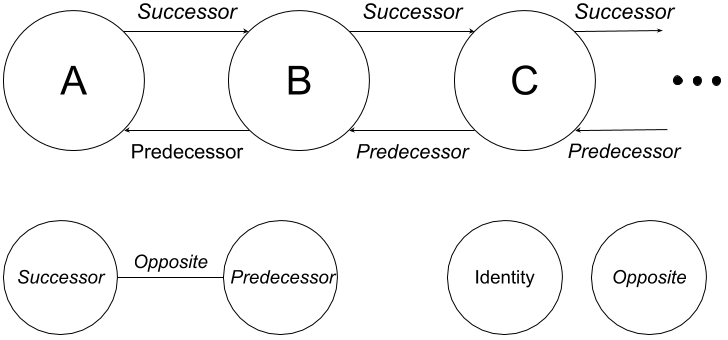}
}{%
    \caption{Part of Slipnet with examples of node length attached to a node, length shrinks if the corresponding node activates.}
    \label{fig:slipnet}
}
\end{floatrow}
\end{figure}

\subsection{Copycat}
Three important components that enable fluidity and creativity are Workspace, Coderack, and Slipnet. The method is described in~\cite{Mitchell1993Analogy-makingPerception}, but  is included for completeness of this paper. In summary, Slipnet enables fluidity through conceptual slippages, Workspace creates a mental blackboard for storing perceived relationships between letters, and Coderack runs Codelets to explore possible outcomes Workspace. Practical implementation methods are discussed in section~\ref{Methods}, and a reinterpretation using modern methedologies of reinforcement learning is described in section~\ref{Reinterpretation}.

\textbf{Workspace} consists of a set of permanent structures for the problem description and a set of temporary structures for working towards a sensible answer. It corresponds to a mental representation of the real world. The structures of Workspace consists of are as follows: letter, group, description, bond, correspondence, rule, and replacement. The Workspace stores each of them in an array or vector. 
\begin{description}[font=$\bullet$~\textit]
\item [Letters and Groups] make up object.
\item [Letter] consists of a single letter.
\item [Group] consists of multiple consecutive letters.
\item [Descriptions, bonds, correspondences, rules, and replacements] make up objects.
\item [Descriptions] are attached to a particular object.
\item [Bonds] are object relationships within initial, modified, target, or answer.
\item [Correspondences] are object relationships between initial, modified, target, or answer.
\item [Rule] is a hypothesis regarding how the initial is transformed to the modified string. 
\item [Replacement] is a rule that applies on the target to the answer string.
\end{description}
    
\textbf{Slipnet} contains platonic concepts that are linked together (see figure~\ref{fig:slipnet}). Nodes that represents concepts are activated when perceived from Workspace. These slippages are giving context of what is relevant to inform Codelet's non-determistic search trajectory. Activated nodes and spread their activations to neighbor nodes who are close. A node is discontinuously fully active when it is activated more than 50\%. Node activation also decays of the activation of a node is directly proportional to conceptual depth --- measuring how deep a pattern is hidden from a perceivable surface symbol.

\textbf{Coderack} consists of an array of codelets instances. A codelet type can be activated by other codelets or activations from the Slipnet. It incorporates ideas from Slipnet that are activated to test plausibility of its hypothesis from Workspace. Currently, codelet instances cooperate in the chained order of scouting-testing-building and fights other incompatible structures. Codelets can be classified into different groups that corresponds to what type of structure it deals with, with the exception of breaker codelet. Random codelet scouts are posted according to the state of Slipnet and Workspace, which finds plausible structure that could be applied. Strength-tester codelets will then be posted by scouts to evaluate how well does the structure fit in context of Workspace and Slipnet state. A builder codelet will be eventually posted by strength-testers to generate the structures in Workspace. During these processes, corresponding concepts in Slipnet may be activated to inform related scouts to be posted. In implementation, a codelet is passed around as lambda-function and gets instantiated when arguments applies to it.
\begin{description}[font=$\bullet$~\textit]
\item Scout codelet probabilistically chooses an object or objects on which to build the structure, and asks "is there any reason for building this type of structure with these objects?"
\item If yes, a strength-tester codelet asks "is the proposed structure strong enough?"
\item If yes, a builder codelet tries to build the structure, fighting against competitors if necessary.
\end{description}

\begin{figure*}[htb]
\centering
    \includegraphics[height=3in, width=3.8in]{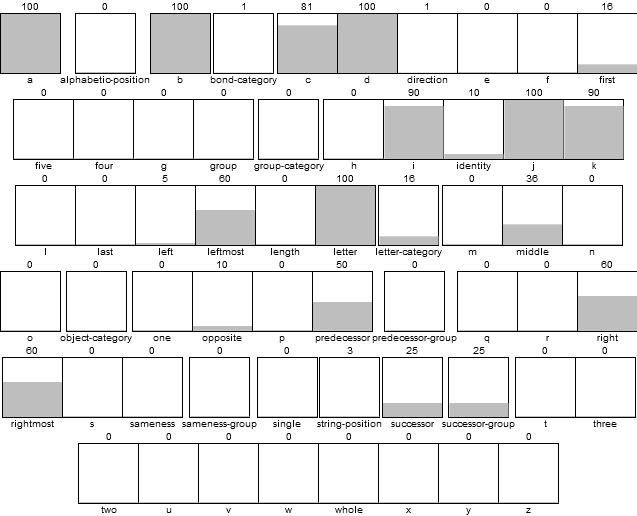}
    \caption{A runtime example of problem abc : abd :: ijk : ?'s Slipnet with activations of node displayed, not including links between nodes.}
\end{figure*}

\textbf{Statistical Regulation}
Coderack also incorporates information --- how deep a concept is (conceptual depth), how well an information is integrated (happiness), how important an information is (importance) --- to inform the probabilities which information should be used in testing the theory it corresponds to. In general, it uses an mechanism of temperature to gauge how deterministic the program should be. It intends to address the \textit{exploration vs. exploitation} problem in the face of many distinct possible answers~\cite{Holland1975AdaptationSystems}. Specifically, codelets measure salience of relevant objects through a function of happiness and importance to make a probabilistic choice. Higher salience will correspond to a higher probability to be used.

\subsection{DrRacket}
We implemented the model in DrRacket. Like most lisp languages, DrRacket is designed to be used functionally. Most syntax other than the low level functions does not use mutations to boost performance and encourage coders to be able to test as there should be no global mutable variables present~\cite{Flatt2018TheV6.12}. In addition, functions can be stored as data to highly abstract. Both factors enable a programmer to understand and write a segment of code quickly.

\begin{lstlisting}[language=Lisp]
#lang racket
(define X 10) ; defines a constant of 10
;myadd1 : number -> number (a functional contract)
(define (myadd1 x) (+ x 1)) ; define function
(check-expect (myadd1 X) 11) ; tests myadd1
\end{lstlisting}

In the context of implementing a complex program such as Copycat, an test or example of the function should be written first for future readers to understand what the function does. The function of the code is an hypothesis of how a programmer can implement according to the test. Following is a pseudo-code example of such philosophy. For large projects of implementations, contracts and tests are kept in a separate runnable file and then copied into a documentation catalog. 

\begin{lstlisting}[language=Lisp]
;update : world-state -> world-state
(define (update state) <code-here>) 
(check-expect (update <old-state>) <new-state>)
\end{lstlisting}

A data structure is defined and used as follows:

\begin{lstlisting}[language=Lisp]
(define-struct mylist (first rest))
(define example1-mylist (make-mylist 1 (make-mylist 2 empty)))
(check-expect (mylist-first example1-mylist) 1)

\end{lstlisting}
\subsection{Function as Arguments}
A function can be passed in as an argument into another function. We used this feature to sort through complex data structures in Copycat and abstract operations. For example, the following built in function apply and map takes in function and a list.

\begin{lstlisting}[language=Lisp]
(check-expect (apply + (list 1 2 3 4 5)) 15)
(check-expect (apply * (list 1 2 3 4 5)) 120)
(check-expect (map add1 (list 1 2 3 4 5)) (list 2 3 4 5 6))
\end{lstlisting}

\subsection{Macros for Making Domain Specific Language}
Macros are preprocessors that transpiles one set of syntax to another. DrRacket itself is built this way that translates \#lang racket to a core set of performance effective instructions~\cite{Felleisen2015TheRM}. The design originated from a set of questions related to how to pass functions as arguments. Eventually, the designers decided to allow translation between datum and syntax, also exposing the macros library. Using this to advantage, one can build another level up from \#lang racket and make an customized language. Details and examples are described in section~\ref{DSL}.

\section{Methods} \label{Methods}
\subsection{Naming Conventions and Memory Allocation}
Both OOP and functional language associates the data with the function. The common perception is that OOP has an advantage of inheritance and a one time memory allocation. One can avoid the problem of hard to keep track function names of different data type by proper naming technique. We named all functions with the data type it operates and the operation: "<data-type>-<operation-name>". For example 
"workspace-structure-to-pointer : workspace structure -> pointer" and "workspace-object-bonds : workspace object -> (list-of-bond)" makes clear of the data it needs and what it does.

Memory allocation problem can be solved by a pointer. Surprisingly, it is not necessary to use traditional memory pointers. Instead, usage of an unique identifier when available, or, a functional pointer that stores how can it reach the data should be preferred. These practices prohibit memory leaks and overflows. The following example 1 and 2 is for Slipnet and Workspace, respectively representing the previous two usage cases. By referencing what nodes are linked through a unique symbol (hashed-string in Racket), there is no need to have an actual node in the link. We later demonstrate that the node reference checking could be completed during compile-time rather in run-time using macros. Similarly, workspace groups consist pointers to letters. Structures of pointers in example 2 can be reached via 
\newline
\lstset{
  aboveskip=0 \medskipamount,
  belowskip=0 \medskipamount
}
\begin{lstlisting}[language=Lisp]
(vector-ref ((pointer-target <a-pointer>) <data-structure>) (pointer-index <a-pointer>)).
\end{lstlisting}

%
%
\begin{lstlisting}[language=Lisp, caption=Slipnet References]
(define-struct slipnet (nodes links)) ; nodes and links are lists
(define-struct link (from to type intrinsic length)) ; from, to is a symbol
(define-struct node (name depth activation)) ; name is a symbol
(make-node 'a 10 0) (make-node 'b 10 0)
(make-link 'a 'b 'lateral 'successor 60)
\end{lstlisting}

\begin{lstlisting}[language=Lisp, caption=Workspace Pointers]
(define-struct workspace (initial modified target answer letters groups descriptions bonds 
correspondences rules) #:mutable)
; pointer : function, integer
(define-struct pointer (target index))
; letter : symbol, pointer
(define-struct letter (string pointer))
; group : symbols, pointers, symbol
(define-struct group (type members direction))
(define-struct description (type pointer))
(define-struct bond (type from to direction))
(define-struct correspondence (type from to))

(define example-workspace (make-workspace "abc" "abd" "ijjkkk" "" (vector example-letter-initial-a example-letter-initial-b #f example-letter-initial-c) (vector example-group-initial-abc #f) (vector empty) (vector empty) (vector empty) (vector empty)))

(define example-group-initial-abc
(make-group 'successor-group (list (make-pointer workspace-letters 0) (make-pointer workspace-letters 1) (make-pointer workspace-letters 2)) 'right))

;workspace-pointer-to-structure : workspace pointer -> structure
(define (workspace-pointer-to-structure space ptr)
  (vector-ref ((pointer-target ptr) space) (pointer-index ptr)))

(check-expect (workspace-pointer-to-structure example-workspace  (first (group-members example-group-initial-abc))) example-letter-initial-a)
\end{lstlisting}
\lstset{
  aboveskip=1 \medskipamount,
  belowskip=1 \medskipamount
}
\subsection{Structural Initialization and Testability Comparison}
Using the previously mentioned recursive syntax for abstraction, the initializations for Copycat are more concise, understandable, and testable. OOP is designed to be flexible and quick to develop, but it also permits bad practices for testing without penalties. Specifically, the following practices lead to behavior combinatorial explosion~\cite{Mathur:2008:FST:1816865}. 
\begin{itemize}
\item Global state dependent behaviors
\item Encapsulation, inheritance, abstract and generic classes.
\end{itemize}
Comparing to a pure functional context, we are only concerned with the first scenario as the rest is not applicable. A well designed functional language such as Racket would make programmers feel syntax resistance and face performance penalties when writing state dependent behavior functions~\cite{Flatt2018TheV6.12}. The performance penalties are on the \textit{set!} functions which mutates a global value. The following two functions of list-sum distinguish in both syntax length and performance. Table~\ref{tab:GlobalVLocal} demonstrates the clear performance difference when two implementations are running on a list with length 1,000 and over 1,000,000 iterations, with garbage collection included in CPU time. The garbage collection time is significantly higher because of higher amount of memory use.

\begin{lstlisting}[language=Lisp]
(define (sum-global lst)
	(let ([s 0])
    	(for-each (lambda (i) (set! s (+ i s))) lst) s))

(define (sum-local lst)  (apply + lst))
\end{lstlisting}
\begin{table}[thb]
  \caption{Racket Global vs. Local and Python Execution Time (ms)}
  \label{tab:GlobalVLocal}
  \begin{tabular}{c|c|c|c}
    \toprule
    Type&CPU Time&Real Time&Garbage Collection Time\\
    \midrule
    Global & 20938 & 21001 & 94   \\
    Local  & 15359 & 15547 & 2895 \\
    Python & - & 49900 & - \\
  \bottomrule
\end{tabular}
\end{table}

All the benefits and drawbacks of both functional and object-oriented implementations are magnified when developing complex programs. The following is a case on initializing the description of a letter, seen on figure~\ref{workspaceInit} and appendix code listing~\ref{table:descInit} on page~\pageref{table:descInit}.

Code~\ref{lst:pyDescInit} shows programmers tend to program linearly when implementing cognitive models in Python. When the procedure's length starts to increase, it is unavoidable to use global states. How can an global behavioral error be amplified and still be untraceable in linear programming? Take the first segment of the for loop, and say we made a human error of swapping the order of two lines. This program's behavior will cascade deep into an experiment trial if there are no proper visualization of the architecture to check or proper tests to write. For researchers, implementation errors will result in either confusion, inaccurate results, or even worse, non-reproducible results. By writing tests for smaller chunks of functions, a programmer can avoid a test-case combinatorial explosion for more complex behaviors. In addition to writing less test, a test error will trace back to the exact function to blame --- saving debugging time. Considering context, Copycat's exploration mechanism is non-deterministic to imitate creativity. In this case, a programmer might know the frequencies of answers are wrong, but to write a test for the entire program is a mere impossibility. With the certainty that a model is implemented properly, a researcher can properly debug the problem of cognitive model instead of code. 

On the other hand, the memory usage of functional implementation could be abnormally high compared to an OOP implementation. This is acceptable in complex softwares of cognitive architectures as a trade off of higher memory usage to higher run time performance and more reliable behaviors. This scenario is preferred in an research environment, where accurate behavior is an priority for understanding something unknown instead of performance. 

The case in workspace also applies to Slipnet, in appendix code listing~\ref{table:nodeInit} on page~\pageref{table:nodeInit}. In slipnet number nodes initialization, functional implementation is more testable due to the separation of each code-block's functionality. In OOP style, an exhaustive list of initialization in will make the testing function also an exhaustive list. Using a functional approach, only three functions need to be tested --- description-tester, description-tester-lambda-get, and the actual abstracted description-tester lambda.

\subsection{Program Performance}
Run time comparison test is not conclusive compared to Python, the different graphics and statistical library performances make two programs hard to compare. But in a general benchmark, Racket is faster than Python in run-time (table~\ref{tab:GlobalVLocal}). This is not a surprise as DrRacket is a functional language with Just-In-Time compiler, while most Python platforms is interpreted. 

\subsection{Teamwork Performance}
    The simple convenience offered by a functional language may not be much of an benefit if there is only one programmer. However, huge research projects like Copycat needs teamwork to be completed on time. Teamwork delegation is very clear for the functional paradigm: cognitive architecture designer can focus on designing experiments to run; program designers understand what they are doing then write contracts with tests; programmer can focus on implementing the functions following contracts. Eventually, these groups can meet and receive feedbacks from each other. An more effective understanding of what the program is doing between people should ease the bottleneck on computational implementations of complex cognitive architectures. 

\section{Future Work}
\subsection{Reinterpretation of Copycat}\label{Reinterpretation}
\begin{table}[thb]
    \caption{Hierarchy of Intelligent Systems, adapted from Time Scales of Human Action. These are the factors that should be considered to understand why and how a phenomenon works.~\cite{Newell1990UnifiedCognition}}
    \label{table:hierarchy}
    \begin{tabular}{c|c|c}
        \toprule
            System&Traditional Frameworks&Artificial Intelligence Frameworks\\
        \midrule
            Environmental & \textbf{Evolutionary} (Ecological) & Philosophy of Artificial Life \\\hline
            Cultural & \textbf{Historical} (Socio-cultural evolution) & Individual evolution \\\hline
            Multi-Agent & \textbf{Social Band} (Sociology, Economics) & Integrative Learning \\\hline
            Task & \textbf{Rational Band} (Psychology) & Individual Heuristics
            \\\hline
            \begin{tabular}{cc}
            Unit Task -\\ Deliberate Act
            \end{tabular} & \textbf{Cognitive Band} (Behavioralism) &
            \begin{tabular}{cc}
            Cognitive Architectures;\\ Reinforcement Learning
            \end{tabular} \\\hline
            \begin{tabular}{cc}
            Neural Circuit -\\ Organelle
            \end{tabular} &
            \begin{tabular}{cc}
            \textbf{Biological Band}\\ (Neuroscience, Info. Theory)
            \end{tabular} & Machine Learning \\
        \bottomrule
    \end{tabular}
\end{table}
Copycat primarily concerns the problem of fluid analogy behavior, which is an example of individual heuristics. The scale of Copycat's alike experiments entail architectures larger than a subfield's concern. As a result, interpretations of the methodologies entailed cannot be simply referred as a localized phenomenon, but a result of multiple factors (see table~\ref{table:hierarchy}). Primarily, the problem of exploitation v. exploration has evolved into a larger reinforcement learning field that concerns how to achieve goals via actions~\cite{Sutton:IRL}, meriting a reinterpretation of Copycat's strength and weaknesses.
Despite Copycat's handcrafted codelets and slipnet suffices for a controlled experiment, to generalize it to other domains entails a large practical time consuming problem.

Reinforcement learning is formalized using Markov decision process as an agent-environment interface: with state, reward, and action. Additionally, the policy space is defined as possible actions associated to state and reward, which a reinforcement learning model optimizes. In terms of reinforcement learning's terminology, policy corresponds to codelets, and the slipnets are the state-transition model. It could be possible to learn slipnet from model-based learning, and incorporate rewards as maximizing or minimizing statistical regulations within Copycat --- conceptual depth, happiness, and importance. The Workspace representations however, are hardwired for the letter domain and we are still searching for a method to overcome. This interpretation needs future experimentation to verify.

\subsection{Extensions to Domain Speific Language} \label{DSL}
The ease of testing in pure functional implementation guarantees expected behavior of the architectures, but what about structural initialization? Currently, we are considering to use DrRacket's built in powerful macros to construct a customized language that simplifies and checks the expressions for Copycat's structural initializations. Visualization of structure could be one way, but complex network type of structure could be very confusing even when visualized. Referencing to the language-oriented programming philosophy, this approach does follows its two subsidiary guidelines~\cite{Felleisen2018ALanguage}.
\begin{itemize}
\item Enable creators of a language to enforce its invariants
\item Turn extra-linguistic mechanisms into linguistic constructs
\end{itemize}
The designer of a DSL can enforce all users to deliver code that satisfies proper reference between structures during compilation. Alternatively, one can enforce code quality during run-time environment, but it cannot be directly tested. It is worth to note that writing languages do require a higher level of mindfulness during designing and coding compared to run-time checks. Despite the additional effort, hardening a set of conversions after experiments have been completed embodies Lakatosian's philosophies directly into software. It could provide higher run-time efficiency by removing the need to manually check structures during run time. The example below on Slipnet initialization is a draft example of so following the philosophies mentioned above, extended from Racket School 2018's teaching example~\cite{Felleisen2018RacketProgramming}. In table~\ref{table:DSL} on page~\pageref{table:DSL}, we propose to define and write an example of language \textit{\#lang s-exp slipnet}. The language is used to define, initialize, and check references of nodes in links, which is important if link lengths can be dependent to a node's activation. Using macros, racket's compiler could theoretically transcompile the following to racket code. \textit{\#lang s-exp slipnet} could run \textit{(link a random 50)} and detect a compile-time error of "link: undefined node: random". But the original implementation (on both python and racket's equivalence) would only detect error during run time, given a proper checking mechanism exists. We have also improved the definition of link and nodes to harden proper length checking, when the length between links are labeled (figure~\ref{fig:slipnet}). To unpack what the translation notation means, we use an simple example adapted from racket school 2018~\cite{Felleisen2018RacketProgramming}.

    Some notations easily convert to normal racket language, such as a value definition. We refer to those as definition as variable definition. For clarification, "\#`" turns the rest of the line into a syntax, while {"\#,"} escapes the literal variable to become an actual value of the variable. If no escapes are needed, {"\#'"} will also turn the line into syntax.
\begin{lstlisting}[language=Lisp]
Definition-Vars =
| (define-value id expr)
   => #`(define-syntax id (#%expression expr))
\end{lstlisting}

    Others might not be very straightforward, such as a function. But it can be broken up into the declaration (define) and execution (apply) part. We put the declaration part into the evaluable definition, while the execution becomes an expression of data. Using the racket function \textit{define-syntax}, \textit{define-function} calls a syntax transformation that matches the identifier of the wanted function. If the function to be defined needs to be executed, \textit{function-app} will look up the value of the function-id, checks arity (number of argument), then transforms the lambda corresponding to the function-id. Note that define-function returns a define-syntax that matches function-id at run-time, while the syntax define-function itself is executed at compile time. This is not the only way it could be done. For example, the usage of recursion requires more elaborate mechanisms.
\begin{lstlisting}[language=Lisp]
Definition-Evals = 
| (define-function (id id1 ...) expr)
   => #`(define-syntax id (cons #,arity (lambda (id1 ...) expr)))
Expression = 
| (function-app id expr1 ...)
  => #`((lambda (id1 ...) expr) expr1 ...)
| Variables : Number, Symbol, List, NULL
\end{lstlisting}

The macros define-function is as follows:
\begin{lstlisting}[language=Lisp]
;Syntax -> Syntax
(define-syntax (define-function stx)
  (syntax-parse stx
    [(_ (f:id param:id ...) body:expr)
     (define arity (length (syntax->list #'(param ...))))
     #`(define-syntax f (cons #,arity (lambda (param ...) body)))]))
\end{lstlisting}

\section{Conclusion}
Proper naming, structural reference, testing, and documentation practices in functional language enables cognitive architectures to be more testable, reusable, and understandable. These properties can be used on \textit{Copycat} to ensure software re-usability. Furthermore, cognitive architectures can benefit from broadening the contextual perspective, as no phenomenons are only caused in a vacuum.

\section{Acknowledgement}
The author would like to thank Mr. Daniel Anderson, the Viewpoint School computer science faculty advisor who suggested Copycat is worth to investigate and advised the approaches to do so. His courses on DrRacket are building a strong computer science community in Viewpoint School. Additionally, the author would like to thank Viewpoint School students who participated in the educational project that helped or attempted to understand what Copycat is and how to implement it efficiently: Anthony Pineci, Ben Zebrack, Brandon Frederick, William Parker, Valen Dunn and Artificial Intelligence Class of 2016-2017. Specifically, naming conventions and tests were suggested and partially rewritten with Anthony Pineci. The author would also like to thank Dong He and Cheng He for advising the methods of writing academic papers. Finally, we would like to thank the Racket community for providing through documentations and Dr. Hofstadter's Fluid Analogies Research Group for proposing and advancing the Copycat project. Our contributions are built on these people's hard work.

\bibliography{Mendeley0714.bib}
\bibliographystyle{plain}

\lstset{
  aboveskip=0 \medskipamount,
  belowskip=0 \medskipamount
}

\appendix
\section{Code}
Following are tables for the code snippets analyzed in the paper.
\subsection{Structural Initialization Functional vs. OOP}
\label{table:descInit}
  	\begin{lstlisting}[language=Lisp, caption=DrRacket Functional Description Initialization, label={lst:racketDescInit}]
;string-position : number number -> symbol
(define (string-position n len)
  (cond
    [(> n len) (error "string-position description position must be smaller than string-length")]
    [(= n 0) 'leftmost]
    [(= n (- len 1)) 'rightmost]
    [(= n (/ ( - len 1) 2)) 'middle]
    [else #f]))
    
(check-expect (string-position 0 1) 'leftmost)
(check-expect (string-position 1 1) 'rightmost)
(check-expect (string-position 1 2) 'middle)
(check-expect (string-position 1 5) #f)
(check-error (string-position 5 1))

;workspace-letter-make-descriptions : workspace letter -> description
(define (workspace-letter-make-descriptions space ltr)
  (local ((define ltr-ptr (workspace-structure-to-pointer space ltr))
          (define pos (string-position (pointer-index (letter-pointer ltr)) (string-length ((pointer-target (letter-pointer ltr)) space))))
          (define desc-pos (if (boolean? pos) empty (list (make-description pos ltr-ptr)))))
      
      (append (list (make-description (string->symbol (letter-string ltr)) ltr-ptr) (make-description 'letter ltr-ptr)) desc-pos)))

(check-expect (workspace-letter-make-descriptions example-workspace example-letter-initial-b)
(list (make-description 'b (make-pointer workspace-letters 1)) (make-description 'letter-category
 (make-pointer workspace-letters 1)) (make-description 'middle (make-pointer workspace-letters 1)))
\end{lstlisting}

\newpage
   \begin{lstlisting}[language=Python, caption=Python OOP Description Initialization, label={lst:pyDescInit}]
def add_descriptions(self):
	for string in [self.initial_string, self.modified_string, self.target_string]:
		for letter in string.get_letters():
			desc = Description(self, letter, self.slipnet.plato_object_category, self.slipnet.plato_letter)
			letter.add_description(desc)
			letter.add_description(desc)
            
			# Whoops, the line above is supposed to be down there #
            
			desc = Description(self, letter, self.slipnet.plato_letter_category, self.slipnet.\
            get_plato_letter(letter.name))
            
			#letter.add_description(desc)#

		lmost_letter = string.get_leftmost_letter()
		if string.length > 1:
			rmost_letter = string.get_rightmost_letter()
			desc = Description(self, lmost_letter, self.slipnet.plato_str_pos_category, self.slipnet.plato_lmost)
            lmost_letter.add_description(desc)
            desc = Description(self, rmost_letter, self.slipnet.\
            plato_str_pos_category, self.slipnet.plato_rightmost)
            rmost_letter.add_description(desc)
		else:
			desc = Description(self, lmost_letter, self.slipnet.plato_str_pos_category, self.slipnet.plato_single)
            lmost_letter.add_description(desc)

		if string.length == 3:
			middle_letter = string.get_letter(1)
			desc = Description(self, middle_letter, self.slipnet.plato_str_pos_category, self.slipnet.plato_middle)
			middle_letter.add_description(desc)
	\end{lstlisting}

\newpage
\subsection{Node Initialization Functional vs. OOP}
  \label{table:nodeInit}
	\begin{lstlisting}[language=Lisp, caption=DrRacket Slipnet Node Initialization, label={lst:racketNodeInit}]
(define nodes-numbers
	(map (lambda(name) (make-node name 30 0)) numbers))
;Description-tester implemented in codelet, following code segment from codelet.

;description-tester-number : object desc -> boolean
(define (description-tester-number desc)
	(and (group? obj) (= (object-length obj) (symbol->number desc))))

(check-expect (description-tester-number (group '(letter)
(list (pointer workspace-initial 0) (pointer workspace-initial 1)))) 'two) #t)
(check-expect (description-tester-number (letter 'a (pointer workspace-initial 0)) 'one) #t)
(check-expect (description-tester empty 'three (letter 'a (pointer workspace-initial 0))) #f)

;workspace-description-tester-lambda-get : workspace symbol(of desc-node) -> lambda
(define (workspace-description-tester-lambda-get space desc)
  (case desc
    [(one two three four five) (lambda (obj) (description-tester-number obj desc))]
    ;other description testers ...
    ))

;Examples: workspace is empty since object-length should not need any context
(check-expect (workspace-description-tester-lambda-get empty 'two) (lambda (obj) (description-tester-number obj 'two)))

;workspace-description-tester : workspace description-type(category) object -> boolean
(define (workspace-description-tester space desc a-obj)
	((workspace-description-tester-lambda-get space desc) a-obj))
;No need to test, this is a wrapper for the behavior-critical functions above.
  \end{lstlisting}

\newpage
  \begin{lstlisting}[language=Python, caption=Python Slipnet Node Initialization, label={lst:pyNodeInit}]
self.slipnet_numbers = []
node = self.add_node("1", 30)
node.description_tester = (lambda obj: obj.type_name == 'group' and obj.length() == 1)
self.slipnet_numbers.append(node)

assert node.description_tester(exampleLen1Group), "Description-tester example failed!"

node = self.add_node("2", 30)
node.description_tester = (lambda obj: obj.type_name == 'group' and obj.length() == 2)
self.slipnet_numbers.append(node)

assert node.description_tester(exampleLen2Group), "Description-tester example failed!"

node = self.add_node("3", 30)
node.description_tester = (lambda obj: obj.type_name == 'group' and obj.length() == 3)
self.slipnet_numbers.append(node)

assert node.description_tester(exampleLen3Group), "Description-tester example failed!"

node = self.add_node("4", 30)
node.description_tester = (lambda obj: obj.type_name == 'group' and obj.length() == 4)
self.slipnet_numbers.append(node)

assert node.description_tester(exampleLen4Group), "Description-tester example failed!"

node = self.add_node("5", 30)
node.description_tester = (lambda obj: obj.type_name == 'group' and obj.length() == 5)
self.slipnet_numbers.append(node)

assert node.description_tester(exampleLen5Group), "Description-tester example failed!"
\end{lstlisting}
\newpage
\subsection{An Domain Specific Language for Slipnet}
  \label{table:DSL}

\begin{lstlisting}[language=Lisp, caption=\#lang s-exp Slipnet Elements and Syntax Translation]
Program = Slipnet-structure | Definition | Expression

Slipnet-structure = 
| (define-node-properties (id ...))
   => #`(define-struct node #,@(id ...))
| (define-link-properties (id ...))
   => #`(define-struct link #,@(id ...))

Identifiers = 
| node-all => #`(list node-id ...)
| link-all => #`(list link-id ...)

Definition-Vars =
| (define-value id expr)
   => (define-syntax id expr)
| (node id expr ...)
   => (define-syntax node-id (expr ...))
| (link id1 id2 expr ...)
   => (define-syntax link-id1-id2 (expr ...))

Definition-Evals = 
| (define-function (id id1 ...) expr)
   => #`(define-syntax id (cons ,arity (lambda (id1 ...) expr)))
| (define-definition (id id1 ...) (list Definition-Vars ...))
   => #`(define-syntax id (cons ,arity (list (lambda (id1 ...) expr) ...)))
| (define-identifiers id (id1 ...))
   => #`(define-syntax id (cons ,arity (list id1 ...)))

Expression = 
| link-id
  => link-id (Racket) => Variable
| node-id
  => node-id (Racket) => Variable
| (function-app id expr1 ...)
  => ((lambda (id1 ...) expr) expr1 ...)
| (definition-app id ...)
  => #`(define-values (id-Definition-Vars ...)
            (values (lambda (id1 ...) expr) ...))
| (identifier-map lambda id-define-identifiers)
  => #`(define-values (id-Definition-Vars ...)
            (values (lambda (id1 ...) expr) ...)))
| Variable : Number, Symbol, List, NULL
\end{lstlisting}

\newpage
\begin{lstlisting}[language=Lisp, caption=Initialization Example]
#lang s-exp slipnet
;default name and from, to in node and link
(define-node-properties (depth [intrinsic-length empty] [activation 0]))
(define-link-properties (type length/node))
(define-value ALPHABET_NODE_DEPTH 10)
(define-value NUMBER_NODE_DEPTH 30)
(define-identifiers ALPHABET_IDENTIFIERS (a b c d ... x y z))
(define-identifiers NUMBER_IDENTIFIERS (one two three four five))

(define (node-group identifiers depth)
  (identifier-map (lambda (id) (node id depth)) identifiers))

(define (links-group identifiers type length/node)
  (identifier-map (lambda (from to) (link from to type length/node)) identifiers))

(node-group ALPHABET_IDENTIFIERS ALPHABET_NODE_DEPTH)
(node-group NUMBER_IDENTIFIERS NUMBER_NODE_DEPTH)

(node length 60)
(node string-position 70)
...
(node successor 50 60)
(node predecessor 50 60)

(define-definition (links-group-double identifiers type ->node node<-)
  (links-group ALPHABET_IDENTIFIERS type ->node)
  (links-group (reverse ALPHABET_IDENTIFIERS) type node<-))

(definition-app links-group-double ALPHABET_IDENTIFIERS 'lateral successor predecessor)
;Numbers...

(definition-app links-group-double ALPHABET_IDENTIFIERS 'slip successor predecessor)
;Numbers...

(provide node-all link-all)

(link-length a b)
(node-activation length)
(node-depth group-cateogry)

(check-exn #rx"link: undefined node: random"
           (lambda ()
             (convert-syntax-error (link a random 50))))
\end{lstlisting}


\end{document}